# Feature Identification and Matching for Hand Hygiene Pose

Rashmi Bakshi

*School of Electrical and Electronic Engineering*

*Technological University, Dublin, Ireland*

**Abstract**

Three popular feature descriptors of computer vision such as SIFT, SURF, and ORB are compared and evaluated. The number of correct features were extracted and matched for the original hand hygiene pose-"Rub hands palm to palm" image and rotated image. An accuracy score is calculated based on the total number of matches and the correct number of matches that were produced. The experiment demonstrated that ORB algorithm outperforms by giving the high number of correct matches in less amount of time. ORB feature detection technique will be applied over handwashing video recordings for the purpose of feature extraction and hand hygiene pose classification as a future work. OpenCV utilized to apply the algorithms within python scripts.

## 1    Introduction

The aim of this paper is to apply popular feature detection algorithms in computer vision towards hand hygiene pose- 'Rub hands palm to palm'. The hand washing video recordings were collected during the previous study with the help of 30 participants demonstrating all the six steps involved in handwashing process [1]. A random video recording was utilised to segment and extract a frame for hand hygiene pose- "Rub hands palm to palm".

The hands were extracted from the image as foreground pixels segmented from the background based on skin detection algorithms. The image was subject to rotation to 90 degrees to test the accuracy of feature descriptor algorithms. A comparison study was carried out where SIFT, SURF and ORB detectors were tested against the number of correct matches produced for image rotation.

## 2    Image Features

A feature is an attribute or a characteristic property of an object, which is tracked.
Local image features can be defined as a specific pattern, which is unique from its neighbour pixels; it can be an edge, corner or a region. An edge refers to the pixel pattern where an intensity changes abruptly. A corner is the intersection of two or more edges in a local neighbourhood. A region is the closed set of connected points with a similar intensity. These local features are converted into numerical descriptors; provide a powerful tool in computer vision and robotic applications such as image retrieval, object tracking, video mining etc. [2].
The term detector or extractor (interchangeably used) refers to the algorithm that detects these local features for further processing where their contents are described i.e. feature descriptor algorithm.

*Characteristics of an ideal local feature* [3]
1. Distinctiveness: The features should be rich in intensity variations so they can be distinguished for feature matching.
2. Locality: Features should be local to reduce the chances of being occluded as well as to allow simple estimation of geometric and photometric deformations between the two frames with different views.
3. Quantity: the total number of detected features (i.e. features density) should be large to reflect the frames content in a compact form.
4. Accuracy: Features detected should be located accurately with respect to different scales, shapes and pixel locations in a frame.
5. Efficiency: Features should be efficiently identified in a short time that makes them suitable for real-time applications.
6. Repeatability: Two frames of the same object (or scene) with different viewing settings, a high percentage of the detected features from the overlapped visible part should be found in both frames.
7. Invariance: Minimum effect on the extracted features in case of large deformation (scale, rotation, etc.)
8. Robustness: Minimum effect on the extracted features in case of small deformation (noise, blur, compression, etc.)

# 3 Popular Image Feature Descriptors

There are various feature detection algorithms in the field of computer vision. In this paper, a comparison study of three most popular descriptors, Scale Invariant Feature Transform (SIFT), Speed up Robust Feature (SURF) and Oriented FAST and Rotated BRIEF (ORB) is carried out with the help of OpenCV library [4]. SIFT is a feature detector developed by Lowe in 2004, proven to be efficient in object recognition applications but computationally expensive and unsuitable for real-time applications [5]. SURF is an approximation of SIFT, performs faster than SIFT without reducing the quality of the detected points [6].

Rublee et al. proposed ORB an efficient alternative to SIFT or SURF in 2011. ORB is a fusion of FAST key point detector and BRIEF descriptor with many modifications to enhance the performance. It is cost effective and provides better matching performance [7].

Karami et al. has conducted a similar study of performance comparison between SIFT, SURF and ORB matching techniques against noise, rotation, scaling and fish eye distortion, demonstrating that ORB is the fastest algorithm but SIFT performs best in all the scenarios [6].

In this paper, these descriptors are compared against a simple hand hygiene pose- 'Rub hands palm to palm'. The descriptor that extracts and fulfils the characteristics of an ideal feature would be further utilised for extracting robust hand features from hand hygiene video recordings that were captured in the previous study [1]. These extracted features will be further passed on to a machine learning classifier for the appropriate identification and classification of hand hygiene stages as a future work.

# 4 Brute-Force Matching and Ratio Test

The basis of Brut-Force matching in OpenCV is simple. The descriptor of one feature in first image is matched with all other features in second image by calculating the distance. The closest features are returned. The algorithm for ORB detector is illustrated. Similar python scripts were implemented for SIFT and SURF [8].

```
Read an input (query) image and a training image; im=cv2.imread('hand.jpg')
Initiate SIFT/SURF/ORB detector; orb = cv2.ORB_create()
Find the key points and descriptors with the initiated detector for both
images; kp, des = orb.detectAndCompute(im, None)
Create bf matcher object; bf = cv2.BFMatcher(cv2.NORM_HAMMING,
crossCheck=True)
Match descriptors; matches = bf.match(des1,des2)
Sort them in the order of their distance; matches = sorted (matches, key =
lambda x:x.distance)
Conduct ratio test;  Good[];
                    for m in matches:
                        if m.distance<0.7:
                            Good.append([m])

Draw first 10
matches.;im=cv2.drawMatches(im1,kp1,im2,kp2,matches[:10],None,flags=2)
Show the image; plt.imshow(im),plt.show()
Save the image. cv2.imwrite('resultORB.jpg',im)
```

## 5    Sample Images and Results

Hand hygiene video recording captured at the laboratory's sink in the previous study [1] were utilised to extract an image of hand hygiene pose- " Rub hands palm to palm ". Skin colour was determined and hands were extracted based on YCbCr color model [9]. Green sheets were used to have a minimum background information. However, skin coloured floor can be seen in the output image in Figure 5.1.

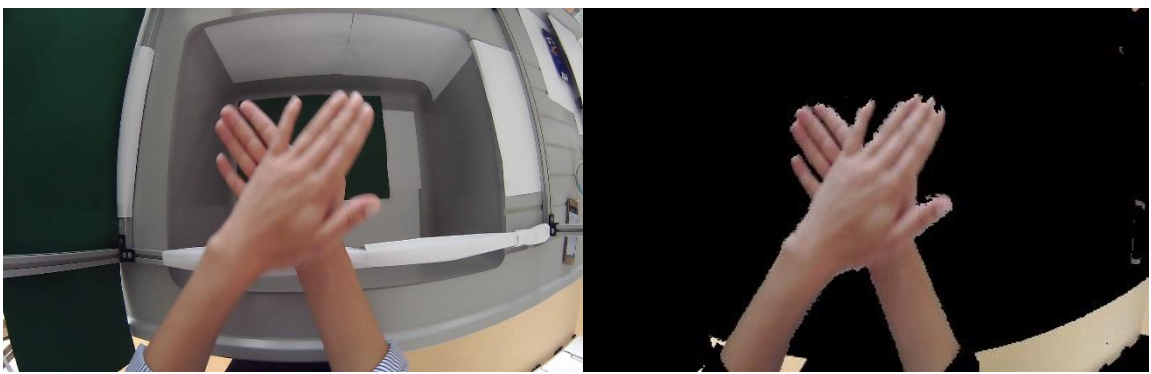

Figure 5.1 Original image; Hands extracted based on YCbCr model [9]

Image rotation by 90 degrees were carried out in order to conduct the feature matching experiments and test the accuracy of the popular feature descriptors such as SIFT, SURF and ORB; their implementation is easily available in OpenCV library. Note: Only first 10 matches are presented in each image. Figure5.2-5.4

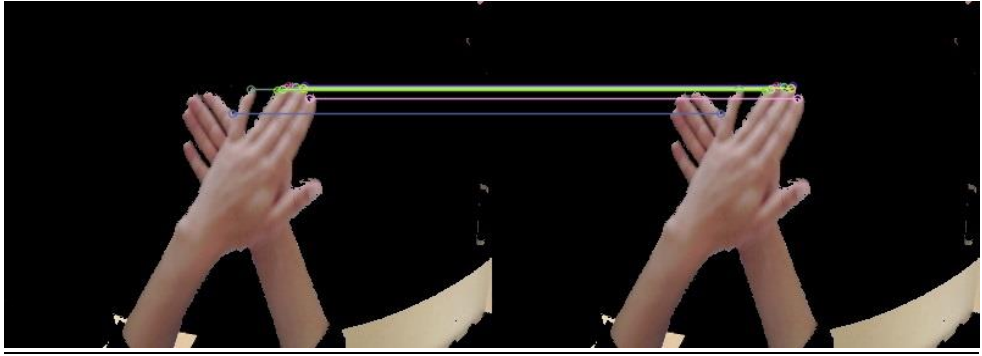

Figure 5.2: Features extracted using ORB for identical images-'Rub hands palm to palm' pose

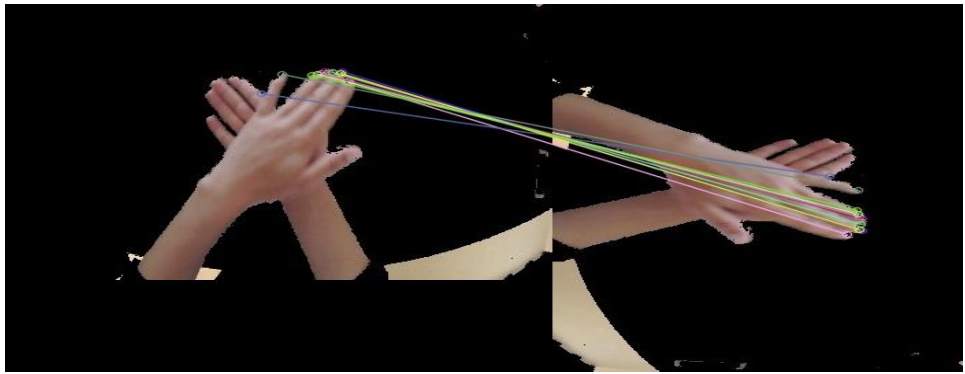

Figure 5.3: Features matching, ORB with original and rotated image (side)

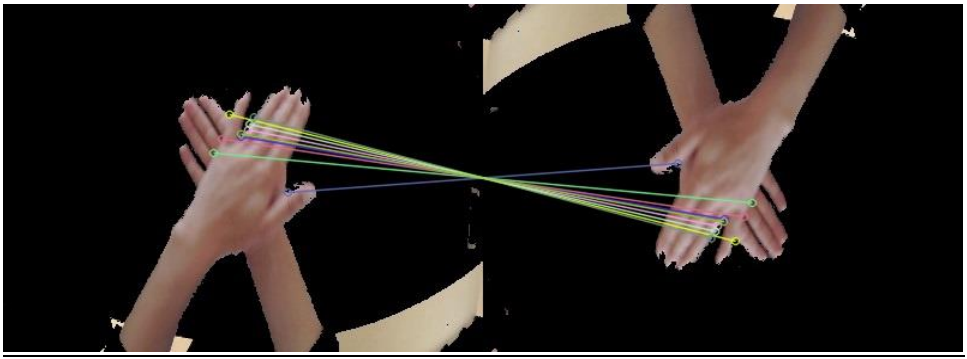

Figure 5.4: Features matching ORB with original and rotated image. First 10 matches shown.

Feature descriptors-SIFT, SURF and ORB descriptors were tested against the original hand image and rotated image (side) for finding the best descriptor for this study that produce increased number of correct matches. Table 5.1 illustrates the calculated accuracy score for each image pair for the mentioned feature descriptors. ORB clearly outperforms SIFT/SURF descriptors with large number of features extracted and 100% accuracy.

| Feature Descriptor | No. of Key points 1 | No. of Key points 2 | Correct no. of matches based on the ratio test m.distance<0.7 | Total No. of Matches | Accuracy score= (Correct matches/Total matches)* 100 % |
|---|---|---|---|---|---|
| SIFT | 109 | 119 | 98 | 109 | 89.9 |
| SURF | 277 | 277 | 277 | 277 | 100 |
| ORB | 433 | 433 | 433 | 433 | 100 |

Table 5.1: Accuracy score is presented for SIFT, SURF and ORB descriptors for hand hygiene pose-'Rub hands palm to palm'.